 \documentclass[tablecaption=bottom,wcp]{jmlr} 

\usepackage{booktabs}
\usepackage{bm}
\DeclareMathOperator*{\argmax}{argmax} 
\usepackage{float}
\usepackage{algpseudocode}
\usepackage{mathtools}
\DeclarePairedDelimiter\ceil{\lceil}{\rceil}

\usepackage[load-configurations=version-1]{siunitx} 

\theorembodyfont{\upshape}
\theoremheaderfont{\scshape}
\theorempostheader{:}
\theoremsep{\newline}

\jmlrproceedings{AABI 2019}{2nd Symposium on Advances in Approximate Bayesian Inference, 2019}

\title[Approximate Inference in Fully Bayesian GPR]{Approximate Inference for Fully Bayesian Gaussian Process Regression}

\author{\Name{Vidhi Lalchand} \Email{vr308@cam.ac.uk} \\ \addr University of Cambridge, Cambridge, UK \\ \addr The Alan Turing Institute, London, UK \\ \\
 \Name{Carl Edward Rasmussen} \Email{cer54@cam.ac.uk} \\ \addr University of Cambridge, Cambridge, UK} 

\usepackage[utf8]{inputenc}

\begin{document}

\maketitle

\begin{abstract}

Learning in Gaussian Process models occurs through the adaptation of hyperparameters of the mean and the covariance function. The classical approach entails maximizing the marginal likelihood yielding fixed point estimates (an approach called \textit{Type II maximum likelihood} or ML-II). An alternative learning procedure is to infer the posterior over hyperparameters in a hierarchical specification of GPs we call \textit{Fully Bayesian Gaussian Process Regression} (GPR). This work considers two approximation schemes for the intractable hyperparameter posterior: 1) Hamiltonian Monte Carlo (HMC) yielding a sampling based approximation and 2) Variational Inference (VI) where the posterior over hyperparameters is approximated by a factorized Gaussian (mean-field) or a full-rank Gaussian accounting for correlations between hyperparameters. We analyse the predictive performance for fully Bayesian GPR on a range of benchmark data sets. 

\end{abstract}

\section{Motivation}
The Gaussian process (GP) posterior is heavily influenced by the choice of the covariance function which needs to be set a priori. Specification of a covariance function and setting the hyperparameters of the chosen covariance family are jointly referred to as the \textit{model selection} problem \citep{rasmussen2004gaussian}. A preponderance of literature on GPs address model selection through maximization of the marginal likelihood, ML-II \citep{mackay1999comparison}. This is an attractive approach as the marginal likelihood is tractable in the case of a Gaussian noise model. Once the point estimate hyperparameters have been selected typically using conjugate gradient methods the posterior distribution over latent function values and hence predictions can be derived in closed form; a compelling property of GP models.\\

While straightforward to implement the non-convexity of the marginal likelihood surface can pose significant challenges for ML-II. The presence of multiple modes can make the process prone to overfitting especially when there are many hyperparameters. Further, weakly identified hyperparameters can manifest in flat ridges in the marginal likelihood surface (where different combinations of hyperparameters give similar marginal likelihood value) \citep{warnes1987problems} making gradient based optimisation extremely sensitive to starting values. Overall, the ML-II point estimates for the hyperparameters are subject to high variability and underestimate prediction uncertainty. \\

The central challenge in extending the Bayesian treatment to hyperparameters in a hierarchical framework is that their posterior is highly intractable; this also renders the predictive posterior intractable. The latter is typically handled numerically by Monte Carlo integration yielding a non-Gaussian predictive posterior; it yields in fact a mixture of GPs. The key question about quantifying uncertainty around covariance hyperparameters is examining how this effect propagates to the posterior predictive distribution under different approximation schemes.

\section{Fully Bayesian GPR}
\label{hgpr}

Given observations $(X, \bm{y}) = \lbrace{ \bm{x_{i}}, y_{i} \rbrace}_{i=1}^{N}$  where $y_{i}$ are noisy realizations of some latent function values $\bm{f}$ corrupted with Gaussian noise, $y_{i} = \bm{f}_{i} + \epsilon_{i}$, $\epsilon_{i} \in \mathcal{N}(0, \sigma_{n}^{2})$, let $k_{\theta}(\bm{x_{i}, \bm{x}_{j}})$ denote a positive definite covariance function parameterized with hyperparameters $\bm{\theta}$ and the corresponding covariance matrix $K_{\theta}$. The hierarchical GP framework is given by,

\begin{align}
\begin{split}
\textrm{Prior over hyperparameters \hspace{5mm} } \bm{\theta} &\sim p(\bm{\theta}) \\
\textrm{Prior over parameters \hspace{2mm} } \bm{f}| X, \bm{\theta} &\sim \mathcal{N}(\bm{0}, K_{\theta}) \\
\textrm{Data likelihood \hspace{5mm}} \bm{y}| \bm{f} &\sim \mathcal{N}(\bm{f}, \sigma_{n}^{2}\mathbb{I})
\label{gen}
\end{split}
\end{align}

The generative model in \eqref{gen} implies the joint posterior over unknowns given as,

\begin{equation}
p(\bm{f}, \bm{\theta} | \bm{y}) = \dfrac{1}{\mathcal{Z}}p(\bm{y}|\bm{f})p(\bm{f}|\bm{\theta})p(\bm{\theta})
\end{equation}

where $\mathcal{Z}$ is the unknown normalization constant. The predictive distribution for unknown test inputs $X^{\star}$ integrates over the joint posterior,

\begin{align}
p(\bm{f}^{\star} | \bm{y}) &= \int \int p(\bm{f}^{\star}| \bm{f}, \bm{\theta})p(\bm{f}, \bm{\theta} |\bm{y})d\bm{f}d\bm{\theta} \\
&= \int \int p(\bm{f}^{\star}| \bm{f}, \bm{\theta})p(\bm{f} | \bm{\theta}, \bm{y})p(\bm{\theta}|\bm{y})d\bm{f}d\bm{\theta}
\end{align}
(where we have suppressed the conditioning over inputs $X, X^{\star}$ for brevity).
The inner integral $\int p(\bm{f}^{\star}| \bm{f}, \bm{\theta})p(\bm{f} | \bm{\theta}, \bm{y})d\bm{f} $ reduces to the standard GP predictive posterior with fixed hyperparameters,

\begin{equation*}
p(\bm{f}^{\star}| \bm{y},\bm{\theta}) = \mathcal{N}(\bm{\mu}^{\star}, \Sigma^{\star})
\label{stan-pred}
\end{equation*}
where,
\begin{align}
\begin{split}
\bm{\mu^{\star}} &=  K_\theta^\star(K_{\theta} + \sigma^{2}_{n}\mathbb{I})^{-1}\bm{y} \hspace{10mm}
\Sigma^{\star} = K_\theta^{\star\star} -  K_\theta^\star(K_{\theta} + \sigma_n^2\mathbb{I})^{-1}K_\theta^{\star^T}
\label{mean-cov}
\end{split}
\end{align}

where $K_{\theta}^{\star\star}$ denotes the covariance matrix evaluated between the test inputs $X^{\star}$ and $K_{\theta}^{*}$ denotes the covariance matrix evaluated between the test inputs $X^{\star}$ and training inputs $X$. 

Under a Gaussian noise setting the hierarchical predictive posterior is reduced to, 

\begin{equation}
p(\bm{f}^{\star} | \bm{y}) = \int p(\bm{f}^{\star}| \bm{y},\bm{\theta})p(\bm{\theta} |\bm{y})d\bm{\theta}  \hspace{3mm} \simeq \hspace{3mm} \dfrac{1}{M}\sum_{j=1}^{M}p(\bm{f^{\star}}| \bm{y}, \bm{\theta_{j}}), \hspace{5mm} \bm{\theta_{j}} \sim p(\bm{\theta} |\bm{y})
\label{pred}
\end{equation}
where $\bm{f}$ is integrated out analytically and $\bm{\theta}_{j}$ are draws from the hyperparameter posterior. The only intractable integral we need to deal with is $p(\bm{\theta}|\bm{y}) \propto p(\bm{y}|\bm{\theta})p(\bm{\theta})$ and predictive posterior follows as per eq. \eqref{pred}. Hence, the hierarchical predictive posterior is a multivariate mixture of Gaussians (Appendix section \ref{ls}).

\section{Methods}
\label{ais}
\subsection{Hamiltonian Monte Carlo (HMC)}

The distinct advantage of HMC over other MCMC methods is the suppression of the random walk behaviour typical of Metropolis and variants. Refer to \citet{neal2011mcmc} for a detailed tutorial. In the experiments we use a self-tuning variant of HMC called the \textit{No-U-Turn-Sampler} (NUTS) proposed in \citet{hoffman2014no} in which the path length is deterministically adjusted for every iteration. Empirically, NUTS is shown to work as well as a hand-tuned HMC. By using NUTS we avoid the overhead in determining good values for the step-size ($\epsilon$) and path length ($L$). We use an identity mass matrix with 500 warm-up iterations and run 4 chains to detect mode switching which can sometimes adversely affect predictions. Further, the primary variables are declared as the log of the hyperparameters $\log(\bm{\theta})$ as this eliminates the positivity constraints that we otherwise we need to account for. The computational cost of the HMC scheme is dominated by the need to invert the covariance matrix $K_{\bm{\theta}}$ which is $\mathcal{O}(N^{3})$.

\subsection{Variational Inference}
\label{advi}

We largely follow the approach in \citet{advi}. We transform the support of hyperparameters $\bm{\theta}$ such that they live in the real space $\mathbb{R}^J$ where $J$ is the number of hyperparameters. Let $\bm{\eta} = g(\bm{\theta}) = \log(\bm{\theta})$ and we proceed by setting the variational family to, $$p(\bm{\eta|y}) \approx q_{\lambda_{mf}}(\bm{\eta}) = \displaystyle\prod_{j=1}^J\mathcal{N}(\eta_{j}|\mu_{j}, \sigma^{2}_{j})$$ in the mean-field approximation where $\lambda_{mf} = (\mu_{1}, \ldots, \mu_{J}, \nu_{1}, \ldots, \nu_{J})$ is the vector of unconstrained variational parameters ($\log(\sigma^{2}_{j}) = \nu_{j}$) which live in $\mathbb{R}^{2J}$. In the full rank approximation the variational family takes the form, $$q_{\lambda_{fr}}(\bm{\eta}) = \mathcal{N}(\bm{\eta}|\bm{\mu}, \bm{LL}^{T})$$ where we use the Cholesky factorization of the covariance matrix $\bm{\Sigma}$ so that the variational parameters $\lambda_{fr} = (\bm{\mu}, \bm{L})$ are unconstrained in $\mathbb{R}^{J+J(J+1)/2}$. The variational objective, ELBO is maximised in the transformed $\bm{\eta}$ space using stochastic gradient ascent and any intractable expectations are approximated using monte carlo integration.  $$ \mathcal{L}(\lambda) = \mathbb{E}_{q_{\lambda}}[\textrm{log}(p(\bm{y}, e^{\eta})) + \textrm{log}|\mathcal{J}_{g^{-1}}(\bm{\eta})|] - \mathbb{E}_{q_{\lambda}}[\textrm{log}(q_{\lambda}(\bm{\eta}))]$$ $$ \lambda^{\star} = \argmax_{\lambda} \mathcal{L}(\lambda)$$ where the term $|\mathcal{J}_{g^{-1}}(\bm{\eta})|$ denotes the Jacobian of the inverse transformation  $g^{-1}(\eta) = e^{\bm{\eta}} =  \bm{\theta}$. The computation of gradients $\nabla_{\bm{\mu}}\mathcal{L}, \nabla_{\bm{\nu}}\mathcal{L}, \nabla_{\bm{L}}\mathcal{L}$ hinges on automatic differentiation and the re-parametrization trick (\citet{kingma2013auto}). The computational cost per iteration is $\mathcal{O}(NMJ)$ where $J$ is the number of hyperparameters and $M$ is the number of MC samples used in computing stochastic gradients. 

\section{Experiments}

We evaluate 4 UCI benchmark regression data sets under fully Bayesian GPR (see Table \ref{results}). For VI we evaluate the mean-field and full-rank approximations. The top line shows the baseline ML-II method. The two metrics shown are: 1) RMSE - square root mean squared error and 2) NLPD - negative log of the predictive density averaged across test data. Except for `wine' which is a near linear dataset, HMC and full-rank variational schemes exceed the performance of ML-II. By looking at Fig.\ref{results} one can notice how the prediction intervals under the full Bayesian schemes capture the true data points. HMC generates a wider span of functions relative to VI (indicated by the uncertainty interval\footnote{see Appendix section \ref{ui} for construction of empirical uncertainty intervals}). The mean-field (MF) performance although inferior to HMC and full-rank (FR) VI still dominates the ML-II method. Further, while HMC is the gold standard and gives a more exact approximation, the VI schemes provide a remarkably close approximation to HMC in terms of error. The higher RMSE of the MF scheme compared to FR and HMC indicates that taking into account correlations between the hyperparameters improves prediction quality. 
\begin{table}[H]
\centering
\scalebox{1}{
\resizebox{\textwidth}{!}{
\begin{tabular}{c|c|c|c|c|c|c|c|c}
Data set & \multicolumn{2}{|c|}{CO$_{2}$} & \multicolumn{2}{|c|}{Wine} & \multicolumn{2} {|c|}{Concrete} &  \multicolumn{2}{|c}{Airline}  \\ \toprule \toprule
Inputs & \multicolumn{2}{|c|}{$N=732, d = 1$} & \multicolumn{2}{|c|}{$N=1599, d = 11$} & \multicolumn{2} {|c|}{$N= 1030, d = 8$} &  \multicolumn{2}{|c}{$N=144, d = 1$}  \\
\toprule
Hyperparameters & \multicolumn{2}{|c|}{$\theta$ = 11} & \multicolumn{2}{|c|}{$\theta$ = 13} & \multicolumn{2} {|c|}{$\theta$ = 10} &  \multicolumn{2}{|c}{$\theta$ = 6} \\
\toprule 
Inference Scheme & RMSE & NLPD & RMSE & NLPD & RMSE & NLPD  & RMSE & NLPD \\
\midrule
ML-II & 4.230 (0.18) & 3.03  & 0.65 (0.02) & 0.98 & 6.12 (0.39) & 3.19 & 21.08 (2.64) & 4.62  \\
HMC (NUTS) & 2.37 (0.10) & 2.53   & 0.65 (0.02) & 0.97 & 5.47 (0.38) & 3.06 & 16.47 (2.34) & 4.31 \\
Mean-field VI &  2.74 (0.12) & 2.05 & 0.65 (0.02) &  0.97 & 5.55 (0.38) & 3.07 & 16.86 (2.49)&  4.36 \\
Full Rank VI & 2.56 (0.12)  & 1.99 & 0.64 (0.02) & 0.97  & 5.52 (0.35) & 3.17 & 16.78 (2.47) & 4.34 \\
\end{tabular}}}
\caption{\small{A comparison of approximate inference schemes for fully Bayesian GPR. For both metrics lower is better, the value in parenthesis denotes standard error of the RMSE.}}
\label{results}
\end{table}
    \vspace{-6mm}
\begin{figure}[t]
    \centering
    \includegraphics{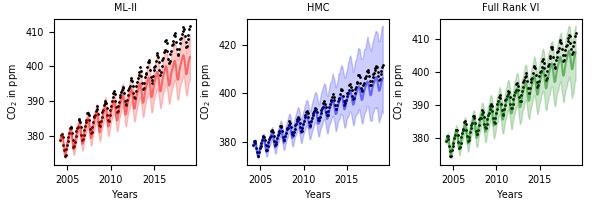}
    \includegraphics{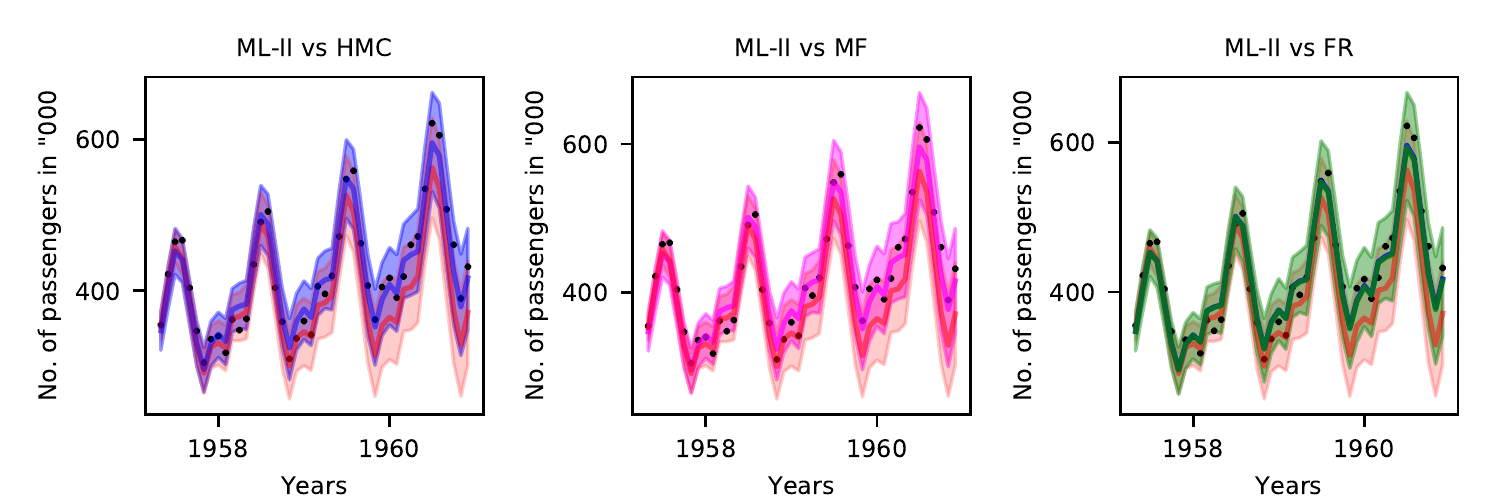}
    \vspace{-4mm}
    \caption{\small{Time-series (test) predictions under Fully Bayesian GPR vs. ML-II (top: CO$_{2}$ and bottom: Airline). In the CO$_{2}$ data where we undertake long-range extrapolation, the uncertainty intervals under the full Bayesian schemes capture the true observations while ML-II underestimates predictive uncertainty. For the Airline dataset, red in each two-way plot denotes ML-II, the uncertainty intervals under the full Bayesian schemes capture the upward trend better than ML-II. The latter also misses on structure that the other schemes capture.}}
    \label{results}
    \vspace{-5mm}
\end{figure}
\vspace{-5mm}
\section{Discussion}

We demonstrate the feasibility of fully Bayesian GPR in the Gaussian likelihood setting for moderate sized high-dimensional data sets with composite kernels. We present a concise comparative analysis across different approximation schemes and find that VI schemes based on the Gaussian variational family are only marginally inferior in terms of predictive performance to the gold standard HMC. While sampling with HMC can be tuned to generate samples from multi-modal posteriors using tempered transitions \citep{neal1996sampling}, the predictions can remain invariant to samples from different hyperparameter modes. Fully Bayesian inference in GPs is highly intractable and one has to consider the trade-off between computational cost, accuracy and robustness of uncertainty intervals. Most interesting real-world applications of GPs entail hand-crafted kernels involving many hyperparameters where there risk of overfitting is not only higher but also hard to detect. A more robust solution is to integrate over the hyperparameters and compute predictive intervals that reflect these uncertainties. An interesting question is whether conducting inference over hierarchies in GPs increases expressivity and representational power by accounting for a more diverse range of models consistent with the data. More specifically, how does it compare to the expressivity of deep GPs \citep{damianou2013deep} with point estimate hyperparameters. Further, these general approximation schemes can be considered in conjunction with different incarnations of GP models where transformations are used to warp the observation space yielding warped GPs \citep{snelson2004warped} or warp the input space either using parametric transformations like neural nets yielding deep kernel learning \citep{wilson2016deep} or non-parametric ones yielding deep GPs \citep{damianou2013deep}. 

\section*{Acknowledgements}
VL is funded by The Alan Turing Institute Doctoral Studentship under the EPSRC grant EP/N510129/1.

\bibliography{main}

\section{Appendix}

\subsection{Related Work}

In early accounts, \citet{neal1998}, \citet{williams1996gaussian} and \citet{barber1997gaussian} explore the integration over covariance hyperparameters using HMC in the regression and classification setting.
More recently, \citet{murray2010slice} use a slice sampling scheme for covariance hyperparameters in a general likelihood setting specifically addressing the coupling between latent function values $\bm{f}$ and hyperparameters $\theta$. \citet{filippone2013comparative} conduct a comparative evaluation of MCMC schemes for the full Bayesian treatment of GP models. Other works like \citet{hensman2015mcmc}  explore the MCMC approach to variationally sparse GPs by using a scheme that jointly samples inducing points and hyperparameters.  \citet{flaxman2015fast} explore a full Bayesian inference framework for regression using HMC but only applies to separable covariance structures together with grid-structured inputs for scalability. On the variational learning side, \citet{snelson2006sparse, titsias2009variational} jointly select inducing points and hyperparameters, hence the posterior over hyperparameters is obtained as a side-effect where the inducing points are the main goal. In more recent work, \citet{yu2019stochastic} propose a novel variational scheme for sparse GPR which extends the Bayesian treatment to hyperparameters.  

\subsection{First and Second moments of the predictive posterior}
\label{ls}
The final form of the hierarchical predictive distribution is a multivariate (location-covariance) mixture of Gaussians:
\begin{equation}
p(\bm{f}^{\star} | \bm{y}) \simeq \frac{1}{M}\sum_{j=1}^{M} \mathcal{N}(\bm{\mu}_{\bm{\theta}_{j}}^{\star}, \Sigma_{\bm{\theta}_{j}}^{\star}) 
\label{h-pred}
\end{equation}

where $\bm{\mu}_{\bm{\theta}_{j}}^{\star}$ and $\Sigma_{\bm{\theta}_{j}}^{\star}$ denote the GP predictive mean and covariance computed with hyperparameter $\bm{\theta_{j}}$. From standard results on Gaussian mixtures we can derive the first and second moments of the hierarchical predictive distribution in \eqref{pred}:
\begin{equation}
E[\bm{f}^{\star} | \bm{y}] = \bm{\mu}_{m}^{\star} =  \frac{1}{M}\sum_{j=1}^{M} \bm{\mu}_{\bm{\theta}_{j}}^{\star} \hspace{5mm}
E[(\bm{f}^{\star} | \bm{y} - \bm{\mu}_{m}^{\star})^{2}]  = \frac{1}{M}\sum_{j=1}^{M} \Sigma_{\bm{\theta}_{j}}^{\star}  + \frac{1}{M}\sum_{j=1}^{M}(\bm{\mu}_{\bm{\theta}_{j}}^{\star} - \bm{\mu}_{m}^{\star})(\bm{\mu}_{\bm{\theta}_{j}}^{\star} - \bm{\mu}_{m}^{\star})^{T} 
 \label{s-pred}
\end{equation}

\subsection{Construction of confidence regions}
\label{ui}

The hierarchical predictive distribution is a mixture of Gaussians and there is no analytical form for the quantiles of a mixture distribution so we can't use the predictive variance in \eqref{s-pred} per se. We estimate quantiles empirically by simulating samples from the univariate mixture distribution at each test input in $X^{\star}$. \\

\begin{algorithm2e}[H]
\caption{95$\%$ Confidence region for the hierarchical predictive distribution}
\begin{algorithmic}
\State \textbf{Given}:  A vector of test inputs $X^{\star} = (X_{1}^{\star}, \ldots, X_{N^{\star}}^{\star})$
\State \textbf{for each} input $X_{i}^{\star}$ where ${i=1, \ldots, N^{\star}}$:
\State \qquad Draw $T$ samples from the univariate mixture distribution $ \hat{f}_{i}^{\star} \sim \frac{1}{M}\sum_{j=1}^{M}\mathcal{N}(\mu_{\bm{\theta_{j}}}^{\star(i)}, \sigma_{\bm{\theta_{j}}}^{\star(i)})$
\State \qquad Sort the samples in ascending order $\hat{f}_{i(1)}^{\star} \leq \ldots \leq \hat{f}_{i(T)}^{\star} $
\State \qquad  Extract the $2.5^{th}$ percentile $\Rightarrow f_{i(r_{l})}^{\star}$ where $r_{l} = \ceil[\Big] {\frac{2.5}{100} \times T}$
\State \qquad  Extract the $97.5^{th}$ percentile $\Rightarrow f_{i(r_u)}^{\star}$ where $r_{u} = \ceil[\Big] {\frac{97.5}{100} \times T}$
\State \textbf{return} 
\State \qquad $\bm{f}_{r_{l}}^{\star} = \lbrace{ f_{i(r_{l})}^{\star} \rbrace}_{i = 1, \ldots, N^{\star}}$ 
\State \qquad $\bm{f}_{r_{u}}^{\star} = \lbrace{ f_{i(r_{u})}^{\star} \rbrace}_{i = 1, \ldots, N^{\star}}$ 
\end{algorithmic}
\end{algorithm2e}

\subsection{Kernels and Choice of Priors}

 All the four data sets use composite kernels constructed from base kernels. Table \ref{kernels} summarizes the base kernels used and the set of hyperparameters for each kernel. All hyperparameters are given vague $\mathcal{N}(0,3)$ priors in log space. Due to the sparsity of Airline data, several of the hyperparameters were weakly identified and in order to constrain inference to a reasonable range we resorted to a tighter normal prior around the ML-II estimates and Gamma(2, 0.1) priors for the noise hyperparameters. All the experiments were done in python using \texttt{pymc3} \citep{salvatier2016probabilistic}. 
 
 \subsection{Experimental Set-up}
 
 In the case of HMC, 4 chains were run to convergence and one chain was selected to compute predictions. For mean-field and full rank VI, a convergence threshold of 1e-4 was set for the variational parameters, optimisation terminated when all the variational parameters (means and standard deviations) concurrently changed by less than 1e-4. For `wine' and `concrete' data sets we use a random 50/50 training/test split. For `CO$_{2}$' we use the first 545 observations as training and for `Airline' we use the first 100 observations as training.\\
     \vspace{10mm}

\begin{table}[H]
\centering
    \begin{tabular}{c|c|c}
    Symbol & Kernel Form & Hyperparameters \\
\hline \hline
$k_{SE}$   &   $\sigma^2_{f}\exp\left(-\dfrac{(x - x')^2}{2\ell^2}\right)$ &  $\{\sigma^2_{f}, \ell\}$\\
    \hline
     $k_{ARD}$  & $\sigma_{f}^{2}\textrm{exp}\left(-\dfrac{1}{2}\sum_{d=1}^{D}\dfrac{(x_{d} - x_{d}^{\prime})^2}{\ell_{d}^{2}}\right)$ & $\{\sigma^2_{f}, \ell_{1}, \ldots, \ell_{D}\}$\\
   \hline
     $k_{RQ}$   &  $\sigma^2_{f} \left( 1 + \dfrac{(x - x')^2}{2 \alpha \ell^2} \right)^{-\alpha}$ &  $\{\sigma^2_{f}, \ell, \alpha\}$ \\
  \hline
     $k_{Per}$ & $\sigma^2_{f}\exp\left(-\dfrac{2\sin^2(\pi|x - x'|/p)}{\ell^2}\right)$ & $\{\sigma^2_{f}, \ell, p\}$\\
   \hline
     $k_{Noise}$ & $\sigma^2_{n}\mathbb{I}_{xx^{\prime}}$ & $\{\sigma^2_{n}\}$ 
    \end{tabular}
    \caption{Base kernels used in the UCI experiments. $k_{SE}$ denotes the squared exponential kernel,   $k_{ARD}$ denotes the automatic relevance determination kernel (squared exponential over dimensions), $k_{Per}$ denotes the periodic kernel, $k_{RQ}$ denotes the rational quadratic kernel and $k_{Noise}$ denotes the white kernel for stationary noise.}
    \vspace{10mm}
    \label{kernels}
    \centering
    \hspace{10mm}
    \begin{tabular}{c|c}
    Data set & Composite Kernel \\
    \hline \hline \\
        CO$_{2}$ &  $k_{SE} + k_{SE}\times k_{Per} + k_{RQ} + k_{SE} + k_{Noise}$ \\
        \hline \\
        Wine & $k_{ARD} + k_{Noise}$ \\
           \hline \\
        Concrete & $k_{ARD} + k_{Noise}$ \\ 
           \hline \\
        Airline &  $k_{SE}\times k_{Per} + k_{SE} + k_{Noise}$\\
    \end{tabular}
    \caption{Composite kernels used in the UCI experiments}
    \label{comp}
\end{table}
\subsection{Further Results}
\vspace{-2mm}
\subsubsection{CO$_{2}$}
\vspace{-3mm}
\begin{figure}[H]
    \centering
    \includegraphics[scale=0.7]{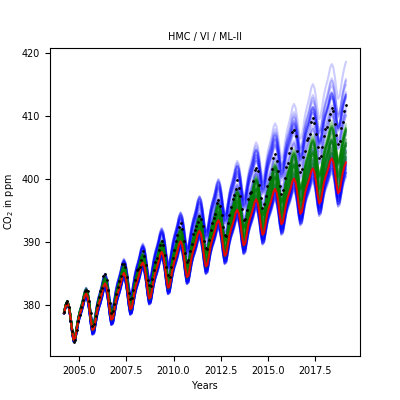}
    \includegraphics[scale=0.4]{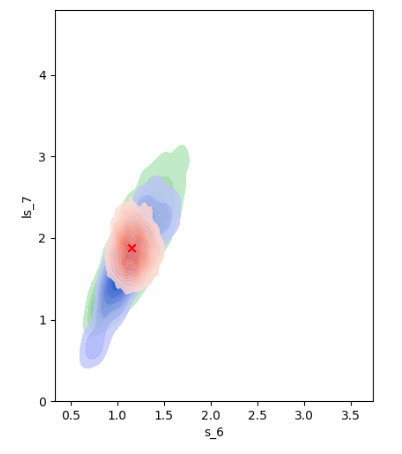}
    \vspace{-3mm}
    \caption{\small{Left: GP means from HMC (blue) and Full Rank VI (green) versus the ML-II GP mean (red). The span of functions tracks the true observations in the long range extrapolation better than ML-II. Right: Bi-variate posterior density between the signal variance and the lengthscale of the $k_{RQ}$ kernel component for the CO$_{2}$ dataset. Blue denotes HMC, green denotes Full Rank VI and orange denotes the mean-field (MF) approximation. MF misses on the structural correlation between the hyperparameters, which is captured by HMC and Full Rank methods.}}
    \label{trio}
\end{figure}

\subsubsection{Airline}

In the figures and tables below, a prefix `s' denotes signal std. deviation, a prefix `ls' denotes lengthscale and a prefix `n' denotes noise std. deviation. The figure below shows marginal posteriors of the hyperparamters used in the Airline kernel. We can make the following remarks: 

\begin{enumerate}
    \item It is evident that sampling and variational optimisation do not converge to the same region of the hyperparameter space as ML-II.
    \item Given that the predictions are better under the full Bayesian schemes, this indicates that ML-II is in an inferior local optimum. 
    \item The mean-field marginal posteriors are narrower than the full rank and HMC posteriors as is expected. Full rank marginal posteriors closely approximate the HMC marginals. 
    \item The noise std. deviation distribution learnt under the full Bayesian schemes is higher than ML-II point estimate indicating overfitting in this particular example.  
\end{enumerate}

\begin{figure}[H]
    \includegraphics[scale=0.5]{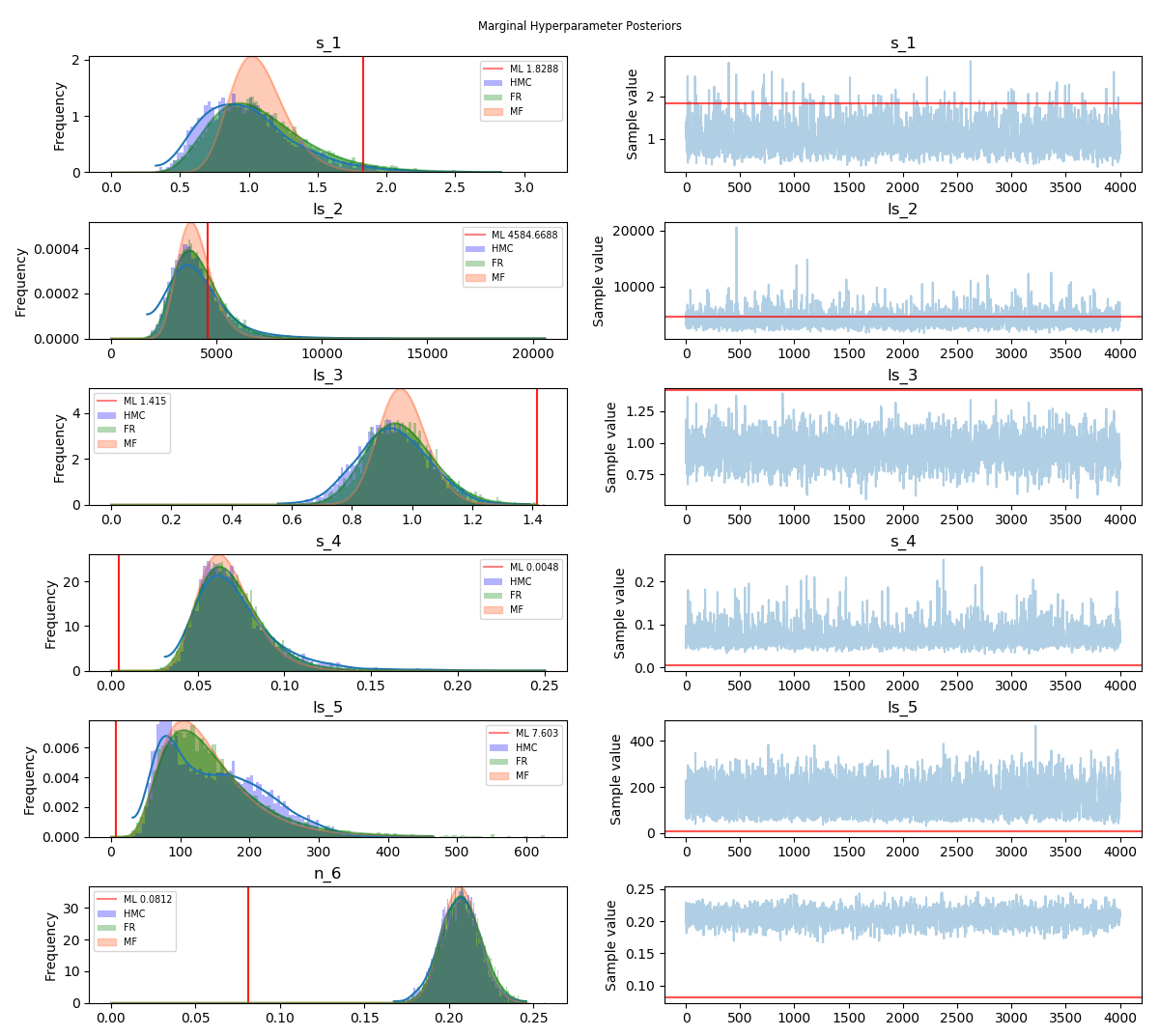}
    \caption{Marginal posteriors under HMC, Mean-Field and Full Rank VI. The vertical red line shows the ML-II point estimate.}
    \label{airline_pos}
\end{figure}

\subsection{Summary of HMC Sampler Statistics}

The tables below summarize statistics based on the trace containing joint samples from the HMC run. The columns hpd$\_$2.5 / hpd$\_$97.5 calculate the highest posterior density interval based on marginal posteriors. n$\_$eff = $\dfrac{MN}{1 + 2 \sum_{t=1}^T \hat{\rho}_t}$ computes effective sample size where $M$ is the number of chains and $N$ is the number of samples in each chain. The numbers below are shown for two chains sampled in parallel with 1000 samples in each chain. $\rho_{t}$ denotes autocorrelation at lag $t$. Rhat denotes the Gelman-Rubin statistic which calculates the ratio of the between chain variance to within chain variance. A Rhat metric close to 1 indicates convergence. 

\subsubsection{CO$_{2}$}
\resizebox{0.8\textwidth}{!}{
\begin{tabular}{c|c|c|c|c|c|c|c}
  
Hyperparameter&mean&sd&mc$\_$error&hpd$\_$2.5&hpd$\_$97.5&n$\_$eff&Rhat\\
\hline \hline
ls$\_$2&103.291&32.318&1.602&51.979&169.806& 624.874 & 0.999\\
ls$\_$4&97.31&25.982&1.618&58.996&148.1& 432.979 & 1.002\\
ls$\_$5&0.802&0.151&0.007&0.542&1.099& 786.430 & 1.003\\
ls$\_$7&1.775&0.585&0.034&0.551&2.832& 916.565 & 0.999\\
ls$\_$10&0.115&0.044&0.002&0.0&0.172&  714.531 & 0.999\\
s$\_$1&224.758&65.185&3.48&124.216&345.636& 882.366 & 0.999 \\
s$\_$3&3.315&1.633&0.094&1.182&6.448& 927.386 & 1.002\\
s$\_$6&1.169&0.307&0.015&0.647&1.702& 724.005 & 1.000\\
s$\_$9&0.155&0.049&0.004&0.0&0.207&  717.402 & 1.008\\
alpha$\_$8&0.121&0.006&0.0&0.11&0.132&  928.689 & 1.002\\
n$\_$11&0.192&0.012&0.001&0.164&0.212&  1021.563& 1.002\\
\end{tabular}}

\subsubsection{Wine}
\resizebox{0.8\textwidth}{!}{
\begin{tabular}{c|c|c|c|c|c|c|c}
Hyperparameter&mean&sd&mc$\_$error&hpd$\_$2.5&hpd$\_$97.5&n$\_$eff&Rhat \\
\hline \hline
s&2.916&0.597&0.035&1.830&3.969 & 835.243 & 1.001\\
ls$\_$0&37.620&44.098&2.604&6.262&110.680 & 474.363 & 1.002\\
ls$\_$1&3.309&1.783&0.087&0.943&6.971 & 936.653 & 1.002\\
ls$\_$2&12.967&19.900&1.008&0.969&39.664 & 725.356 & 1.000\\
ls$\_$3&67.047&66.214&3.627&12.987&155.405 & 645.765 & 0.999\\
ls$\_$4&5.211&10.276&0.585&0.346&21.110 & 853.601 & 0.999\\
ls$\_$5&196.192&275.433&17.662&22.056&607.781 & 936.735 & 0.998\\
ls$\_$6&379.519&224.737&12.508&84.270&821.381 & 1032.174 & 0.999\\
ls$\_$7&3.766&8.182&0.377&0.039&16.234 & 982.004 & 0.998\\
ls$\_$8&10.990&14.306&0.700&1.049&41.657 & 935.461 & 0.999\\
ls$\_$9&1.203&0.568&0.033&0.530&2.448 & 826.143 & 1.003\\
ls$\_$10&4.002&1.890&0.160&2.351&5.565 & 723.359 & 1.004\\
n&0.778&0.010&0.000&0.759&0.797 & 629.475 & 1.000   
\end{tabular}}

\subsubsection{Concrete}
\resizebox{0.8\textwidth}{!}{
\begin{tabular}{c|c|c|c|c|c|c|c}
Hyperparameter&mean&sd&mc$\_$error&hpd$\_$2.5&hpd$\_$97.5&n$\_$eff&Rhat \\
\hline \hline
s&35.714&3.792&0.149&28.585&42.981&581.845&1.000 \\
ls$\_$0&460.767&78.844&2.651&330.721&635.389&924.768&1.005 \\
ls$\_$1&398.286&72.457&2.491&270.638&541.433&845.690&1.000 \\
ls$\_$2&257.044&111.277&4.653&89.867&472.549&610.105&0.999 \\
ls$\_$3&28.162&2.997&0.111&22.473&33.914&676.929&0.999 \\
ls$\_$4&21.019&4.844&0.205&13.091&30.560&528.266&0.999 \\
ls$\_$5&227.006&84.380&4.501&115.147&366.782&310.749&1.000 \\
ls$\_$6&281.485&49.848&1.564&187.606&381.976&949.561&0.999 \\
ls$\_$7&63.033&6.296&0.222&50.671&75.463&834.811&0.999 \\
n&1.959&0.036&0.001&1.884&2.028&707.956&1.003 
\end{tabular}}



\end{document}